\PassOptionsToPackage{pdftex}{graphicx}
\documentclass[sigconf]{acmart}
\usepackage{tikz}
\usepackage{amsmath}
\usetikzlibrary{arrows,positioning} 
\tikzset{
    >=stealth',
    punkt/.style={
           rectangle,
           rounded corners,
           draw=black, very thick,
           text width=6.5em,
           minimum height=2em,
           text centered},
    pil/.style={
           ->,
           thick,
           shorten <=2pt,
           shorten >=2pt,}
}

\usepackage{lipsum}
\usepackage{mathtools}
\usepackage{cuted}
\usepackage{longtable,tabu}
\usepackage{graphicx}

\AtBeginDocument{%
  \providecommand\BibTeX{{%
    \normalfont B\kern-0.5em{\scshape i\kern-0.25em b}\kern-0.8em\TeX}}}

\copyrightyear{2020}
\acmYear{2020}
\setcopyright{rightsretained}
\acmConference[KDD '20]{Proceedings of the 26th ACM SIGKDD Conference on Knowledge Discovery and Data Mining}{August 23--27, 2020}{Virtual Event, CA, USA}
\acmBooktitle{Proceedings of the 26th ACM SIGKDD Conference on Knowledge Discovery and Data Mining (KDD '20), August 23--27, 2020, Virtual Event, CA, USA}
\acmDOI{10.1145/3394486.3403206}
\acmISBN{978-1-4503-7998-4/20/08}
\newcommand{\littleheader}[1]{\noindent \textbf{#1}}

\begin{document}
\fancyhead{}

\title{Deep State-Space Generative Model For Correlated Time-to-Event Predictions}

\author{Yuan Xue}
\email{yuanxue@google.com}
\affiliation{%
  \institution{Google Inc.}
  \country{USA}
}

\author{Denny Zhou}
\email{dennyzhou@google.com}
\affiliation{%
  \institution{Google Inc.}
  \country{USA}
}

\author{Nan Du}
\email{dunan@google.com}
\affiliation{%
  \institution{Google Inc.}
  \country{USA}
}

\author{Andrew M. Dai}
\email{adai@google.com}
\affiliation{%
  \institution{Google Inc.}
  \country{USA}
}

\author{Zhen Xu}
\email{zhenxu@google.com}
\affiliation{%
  \institution{Google Inc.}
  \country{USA}
}

\author{Kun Zhang}
\email{kunzhang@google.com}
\affiliation{%
  \institution{Google Inc.}
  \country{USA}
}

\author{Claire Cui}
\email{claire@google.com}
\affiliation{%
  \institution{Google Inc.}
  \country{USA}
}

\renewcommand{\shortauthors}{Xue and Zhou, et al.}

\begin{abstract}
Capturing the inter-dependencies among multiple types of clinically-critical events is critical not only to accurate future event prediction, but also to better treatment planning. In this work, we propose a deep latent state-space generative model to capture the interactions among different types of correlated clinical events (e.g., kidney failure, mortality) by explicitly modeling the temporal dynamics of patients' latent states. Based on these learned patient states, we further develop a new general discrete-time formulation of the hazard rate function to estimate the survival distribution of patients with significantly improved accuracy. Extensive evaluations over real EMR data show that our proposed model compares favorably to various state-of-the-art baselines. Furthermore, our method also uncovers meaningful insights about the latent correlations among mortality and different types of organ failures. 

\end{abstract}



\maketitle

% tikzlibrary.code.tex
%
% Copyright 2010-2011 by Laura Dietz
% Copyright 2012 by Jaakko Luttinen
%
% This file may be distributed and/or modified
%
% 1. under the LaTeX Project Public License and/or
% 2. under the GNU General Public License.
%
% See the files LICENSE_LPPL and LICENSE_GPL for more details.

% Load other libraries
\usetikzlibrary{shapes}
\usetikzlibrary{fit}
\usetikzlibrary{chains}
\usetikzlibrary{arrows}

% Latent node
\tikzstyle{latent} = [circle,fill=white,draw=black,inner sep=1pt,
minimum size=20pt, font=\fontsize{10}{10}\selectfont, node distance=1]
% Observed node
\tikzstyle{obs} = [latent,fill=gray!25]
% Constant node
\tikzstyle{const} = [rectangle, inner sep=0pt, node distance=1]
% Factor node
\tikzstyle{factor} = [rectangle, fill=black,minimum size=5pt, inner
sep=0pt, node distance=0.4]
% Deterministic node
\tikzstyle{det} = [latent, diamond]

% Plate node
\tikzstyle{plate} = [draw, rectangle, rounded corners, fit=#1]
% Invisible wrapper node
\tikzstyle{wrap} = [inner sep=0pt, fit=#1]
% Gate
\tikzstyle{gate} = [draw, rectangle, dashed, fit=#1]

% Caption node
\tikzstyle{caption} = [font=\footnotesize, node distance=0] %
\tikzstyle{plate caption} = [caption, node distance=0, inner sep=0pt,
below left=5pt and 0pt of #1.south east] %
\tikzstyle{factor caption} = [caption] %
\tikzstyle{every label} += [caption] %

\tikzset{>={triangle 45}}

%\pgfdeclarelayer{b}
%\pgfdeclarelayer{f}
%\pgfsetlayers{b,main,f}

% \factoredge [options] {inputs} {factors} {outputs}
\newcommand{\factoredge}[4][]{ %
  % Connect all nodes #2 to all nodes #4 via all factors #3.
  \foreach \f in {#3} { %
    \foreach \x in {#2} { %
      \path (\x) edge[-,#1] (\f) ; %
      %\draw[-,#1] (\x) edge[-] (\f) ; %
    } ;
    \foreach \y in {#4} { %
      \path (\f) edge[->,#1] (\y) ; %
      %\draw[->,#1] (\f) -- (\y) ; %
    } ;
  } ;
}

% \edge [options] {inputs} {outputs}
\newcommand{\edge}[3][]{ %
  % Connect all nodes #2 to all nodes #3.
  \foreach \x in {#2} { %
    \foreach \y in {#3} { %
      \path (\x) edge [->,#1] (\y) ;%
      %\draw[->,#1] (\x) -- (\y) ;%
    } ;
  } ;
}

% \factor [options] {name} {caption} {inputs} {outputs}
\newcommand{\factor}[5][]{ %
  % Draw the factor node. Use alias to allow empty names.
  \node[factor, label={[name=#2-caption]#3}, name=#2, #1,
  alias=#2-alias] {} ; %
  % Connect all inputs to outputs via this factor
  \factoredge {#4} {#2-alias} {#5} ; %
}

% \plate [options] {name} {fitlist} {caption}
\newcommand{\plate}[4][]{ %
  \node[wrap=#3] (#2-wrap) {}; %
  \node[plate caption=#2-wrap] (#2-caption) {#4}; %
  \node[plate=(#2-wrap)(#2-caption), #1] (#2) {}; %
}

% \gate [options] {name} {fitlist} {inputs}
\newcommand{\gate}[4][]{ %
  \node[gate=#3, name=#2, #1, alias=#2-alias] {}; %
  \foreach \x in {#4} { %
    \draw [-*,thick] (\x) -- (#2-alias); %
  } ;%
}

% \vgate {name} {fitlist-left} {caption-left} {fitlist-right}
% {caption-right} {inputs}
\newcommand{\vgate}[6]{ %
  % Wrap the left and right parts
  \node[wrap=#2] (#1-left) {}; %
  \node[wrap=#4] (#1-right) {}; %
  % Draw the gate
  \node[gate=(#1-left)(#1-right)] (#1) {}; %
  % Add captions
  \node[caption, below left=of #1.north ] (#1-left-caption)
  {#3}; %
  \node[caption, below right=of #1.north ] (#1-right-caption)
  {#5}; %
  % Draw middle separation
  \draw [-, dashed] (#1.north) -- (#1.south); %
  % Draw inputs
  \foreach \x in {#6} { %
    \draw [-*,thick] (\x) -- (#1); %
  } ;%
}

% \hgate {name} {fitlist-top} {caption-top} {fitlist-bottom}
% {caption-bottom} {inputs}
\newcommand{\hgate}[6]{ %
  % Wrap the left and right parts
  \node[wrap=#2] (#1-top) {}; %
  \node[wrap=#4] (#1-bottom) {}; %
  % Draw the gate
  \node[gate=(#1-top)(#1-bottom)] (#1) {}; %
  % Add captions
  \node[caption, above right=of #1.west ] (#1-top-caption)
  {#3}; %
  \node[caption, below right=of #1.west ] (#1-bottom-caption)
  {#5}; %
  % Draw middle separation
  \draw [-, dashed] (#1.west) -- (#1.east); %
  % Draw inputs
  \foreach \x in {#6} { %
    \draw [-*,thick] (\x) -- (#1); %
  } ;%
}

\section{Introduction}

Time-to-event prediction (also known as survival analysis) investigates the distribution of time duration until the event of interest happens in the presence of event censorship. In the healthcare domain, it is an essential tool for modeling the risks of critical medical events and capturing of the relationship between the co-variants and the risks~\cite{cox}.

Recently, machine learning methods have been applied to time-to-event predictions to provide flexible modeling of the time distribution~\cite{aaai18-schaar-deephit, bmc18-katzman-deepsurv, icml-date}, and capture the nonlinear relationship between co-variants and the risk of an event~\cite{pmlr16-blei-dsa}. Most of the prior work~\cite{icml-date, bmc18-katzman-deepsurv, aaai19-ren} on time-to-event prediction are limited in modeling a single type of event, and lack the capability of analyzing the correlations among the risks of multiple types of events. In real world, most events are by nature related to or even caused by one another. In particular, within the medical domain, death may be caused by either a single organ failure or a combination of multiple simultaneous organ failures which could significantly increase the risk of death in a non-linear fashion. Furthermore, the dysfunction or failure of one organ will also trigger the dysfunction/failure of another (e.g., kidney failure may be caused by liver damage). Therefore, predicting the next-occurrence of events of interest (e.g., death) heavily depends on the joint risks of other associated types of events (e.g., different organ failures). 

Understanding and capturing the inter-dependencies among multiple types of clinically-critical events are important not only to deriving more accurate future event timing predictions, but also critical for designing respective treatment plans that could simultaneously handle multiple correlated life-threatening failure events. For instance, when designing optimal treatment plans for patients with comorbidities, the decision on whether a diabetic patient (who also has a renal disease) should receive dialysis or a renal transplant must be based on a joint prognosis of diabetes-related complications and end-stage renal failure. Overlooking the diabetes-related risks may lead to misguided therapeutic decisions. 

Recognizing the necessity of a multi-type event model, a sequence of recent work~\cite{kdd16-li-multitask,icdm17-li-multitask,nips17-schaar-gaussian,aaai18-schaar-deephit,nips18-schaar-boosting,nips18-zhang-lomax} propose multi-type event analysis with a particular parametric form of event relation, \emph{the competing risk}~\cite{fine-gray}, that is, the occurrence of one event precludes the occurrence of another. However, uncovering the general temporal correlations among multiple types of events still remains an open question.
%%% WE argue that the shared state mode..

Recent wide adoption of electronic medical records (EMR) leads to the collection of an enormous amount of patient measurements over time in the form of time-series data. These retrospective data contain valuable information that captures the intricate relations among patient conditions, clinical interventions and outcomes, and present a promising avenue for accurately capturing the temporal progression of clinically critical events.

To accurately predict the temporal progression and establish the temporal correlations of multiple types of a patient's critical medical events, we need a powerful model that is able to capture the dynamics of the underlying patient states from rich time-varying patient measurements and infer the inter-dependency of the occurrences of clinically critical events from these hidden dynamic states.

Therefore, in this work, we present a deep state-space generative model, which provides joint time-to-event predictions of multiple clinical events based on the EMR time-series data. Our model is able to simultaneously predict mortality risk and capture organ failure risk trajectories by leveraging the temporal progressive correlations between the past measurements and clinical interventions. More specifically, we have made the following contributions:

\littleheader{Technical Significance}. We present a deep state space generative model, augmented with intervention forecasting, to provide a principle framework to capture the interactions among observations, interventions, critical event occurrences, latent patient states and their uncertainty. Based on the temporal dynamics of patients' states, we develop a new discrete-time hazard rate model that provides flexible fitting of general time-to-event distributions without restricted parametric assumptions.

\littleheader{Clinical Value}. The ability to jointly forecast multiple clinical events and identify their temporal correlations provides clinicians with a full picture of a patient’s medical condition and better supports them with decision making. Moreover, by demonstrating the correlations between the risk of these organ failure events and mortality, we also provide physicians with evidence to understand our mortality risk predictions.

\section{Related Work}
This section mainly summarizes the related studies in literature as follows:

\textbf{Clinical predictions}. Deep learning models are increasingly used to improve the predictions of clinical outcomes, such as mortality or diagnosis~\cite{rajkomar18scalable,bcb2017-sha,Che2018-vj,Choi2015-ak,iclr16-lipton}. These studies can be roughly categorized based on the data, models and the prediction tasks used. Our work uses clinical time-series data, similar to ~\cite{aaai2018-song,aim2013-liu,aaai2016-liu,iclr16-lipton,jaim2017-wu} to make predictions. It is related to the switching discrete state space model\cite{ghassemi-switching} used to predict clinical interventions in ICU, and the continuous state space model\cite{pmlr-v68-raghu17a}, which learns a treatment policy using deep reinforcement learning. However, our task, providing a joint forecast for the hazard rate of multiple correlated event, has not been considered by these previous work.

\textbf{Time-to-event predictions}. Machine learning methods have been applied to time-to-event predictions. For example, recent works have extended the classical Cox proportional hazards model with neural network-based co-variate encoding~\cite{bmc18-katzman-deepsurv,icann-rnn-surv} and with multi-task formulations~\cite{icdm17-li-multitask,kdd16-li-multitask}. The work of~\cite{aaai19-ren} converts the time-to-event estimation to a discretized-time classification problem, while others use a continuous-time model based on Gaussian processes~\cite{nips16-fernandez-gaussian,nips17-schaar-gaussian,sim10-barrett-gaussian} or generative adversarial networks~\cite{icml-date} to model the nonlinear relationship between co-variates and the time. Unlike existing works, our work combines a deep state space model with a discrete-time hazard model to support more flexible distribution model. The predicted hazard rates of correlated events are calibrated in time, which provide better interpretations on the influence among the events.

\textbf{Deep state-space models}. A few recent works~\cite{nipsw15-dkf,aaai17-sontag-dmm,iclr17-dvbf,nips17-marco-kvae,nips16-marco-srnn,nips19-schaar-attentive,nipsw19-xue-dssm} have extended the traditional linear Gaussian state space model to handle non-linear relations via neural networks. The goal of these works is to fit a generative state space model to a sequence of observations and actions, while ours is to capture the inherent state transition dynamics and use it for time-to-event prediction. In particular, we focus on the modeling and learning of hazard rate functions of different events which share the common underlying state.  In addition, these existing works have presented different models, methods and neural network architectures to infer the latent states more accurately. From this perspective, our work is complementary to these existing works. The encoding network in our method can leverage any existing architecture. In the experiment, we adopted the architecture presented in \cite{nipsw15-dkf}.
\newcommand{\pth}{p_\theta}
\newcommand{\qph}{q_\phi}
\newcommand{\xfull}{\mathbf{x}_{1:T}}
\newcommand{\zfull}{\mathbf{z}_{1:T}}
\newcommand{\utwofull}{\mathbf{u}_{2:T}}
\newcommand{\ufull}{\mathbf{u}_{1:T}}
\newcommand{\x}{\mathbf{x}}
\newcommand{\z}{\mathbf{z}}
\newcommand{\sumt}{\sum_{t=1}^T}
\newcommand{\sumtm}{\sum_{t=1}^{T-1}}
\newcommand{\xh}{\mathbf{x}_{1:t^*}}
\newcommand{\zh}{\mathbf{z}_{1:t^*}}
\newcommand{\uh}{\mathbf{u}_{1:t^*}}
\newcommand{\xf}{\mathbf{x}_{t^*+1:t^*+\tau}}
\newcommand{\zf}{\mathbf{z}_{t^*+1:t^*+\tau}}
\newcommand{\uf}{\mathbf{u}_{t^*+1:t^*+\tau}}
\newcommand{\sumf}{\sum_{t=t^*+1}^{t^*+\tau}}
\newcommand{\Exp}[2]{\mathop{\mathbb{E}}_{#2}\left[#1\right]}

\section{Correlated Time-to-Event Predictions}
\label{sec:problem}

\begin{figure*}[!hbtp]
\begin{center}
\includegraphics[scale=0.35]{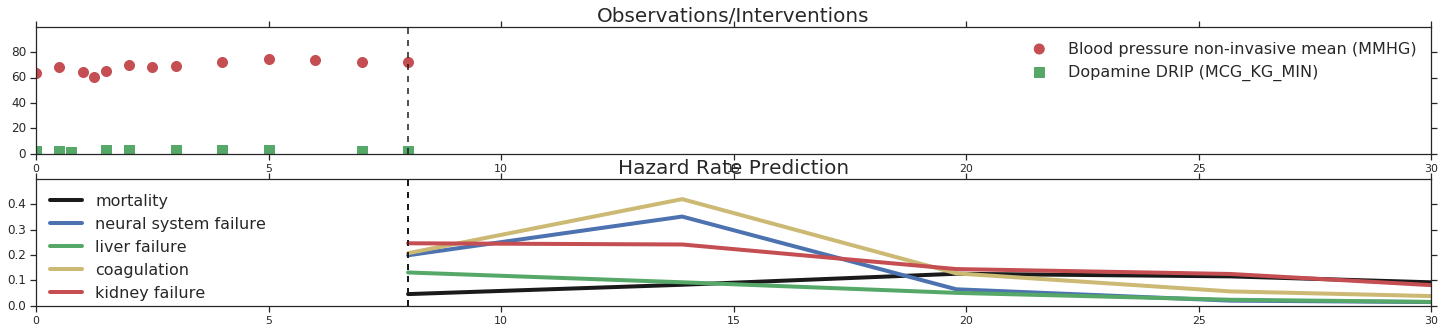}
\end{center}
\caption{Time-calibrated multiple event risk trajectory. Given the past blood pressure readings (red dots) and Dopamine dosage (green square) in the upper left, the model captures how the risk of different types of clinical outcomes (mortality, neural system failure, etc.) can change with time in the future at the bottom right.}
\label{fig:compass}
\end{figure*}

We first describe our usage scenario and problem setting then provide the formal definition of our learning task in this section. Electronic medical record (EMR) provides a longitudinal database where each patient medical history including, lab and vital measurements, medication orders and administrations, conditions and diagnosis, medical procedures, is recorded. We are utilizing the measurements and interventions from EMR in the form of time-series data to predict the time of future critical medical events such as mortality and organ failures. To capture the uncertainty of time, we estimate the distribution of time duration between the time of prediction and the event occurrence as well as the risk of experiencing the events of interest at any given time.

Formally, consider a longitudinal EMR system with $N$ patients. We discretize and calibrate patient $i$'s longitudinal records to a time window $[1, T_i]$, where time $1$ and $T_i$ represent the time when the patient first and last interacts with the system\footnote{For inpatient prediction, this period refers to the start and end of an inpatient encounter, instead of the entire patient history.}. Note that $T_i$, also called censor time in survival analysis, can vary for different patients $i$. In this paper, we focus on personalized predictions. When the context is clear, we simplify notation $T_i$ with $T$. We consider two types of time series data in EMR:
\begin{itemize}
    \item \textbf{Observations} $\mathbf{x}$, a real-valued vector of $O$-dimension. Each dimension corresponds to one type of clinical measurement including vital signs and lab results (e.g., mean blood pressure, serum lactate). We use $\mathbf{x}_{1:T}$ to denote the sequence of measurements at discrete time points $t = 1, ..., T$; 
    \item \textbf{Interventions} $\mathbf{u}$, a real-valued vector of $I$-dimension. Each dimension corresponds to one type of clinical intervention, and its value indicates the presence and the level of intervention such as the dosage of medication being administrated or the settings of a mechanical ventilator. Similarly, $\mathbf{u}_{1:T}$ denotes the sequence of interventions at $t = 1, \ldots, T$. 
\end{itemize}

At prediction time $t^*$, given the sequence of observations and interventions $\mathbf{x}_{1:t^*}$, $\mathbf{u}_{1:t^*}$, we estimate the distribution of time for a set of clinically significant events. We represent an event $e$ with a tuple $(c,t^e)$, where $t^e$ denotes the time to the event from $t^*$ and $c$ is the censorship indicator. If the event is observed, then $t^e \leq T$ and $c=0$; Otherwise the event is censored and $t^e = T$ and $c=1$. 

The time-to-event distribution is well captured by two functions: 
\begin{itemize}
    \item \textit{Survival function} $S^e(t) = \Pr(t^e \geq t)$, a monotonically decreasing function representing the probability of $t^e$ not earlier than $t$;
    \item \textit{Hazard function} $\lambda^e(t)$ representing the instantaneous rate of an event occurrence at time $t$ given that no event occurred before time $t$. 
\end{itemize}

As detailed in Sec.~\ref{sec:model.survival}, $\lambda^e(t)$ determines $S^e(t)$ and captures the instantaneous risk of a patient experiencing event $e$ at $t$. As thus, the time-to-event prediction can be achieved by estimating $\lambda^e(t^*+\tau)$ where $\tau \in [1, H]$, $H$ being the maximum time length of the prediction horizon, for a set of events of interest $e\in E$. 

Our learning task is to estimate the conditional probability distribution $\lambda^e(t^*+\tau |\bar{\x}, \bar{\mathbf{u}})$, where $\bar{\x}, \bar{\mathbf{u}}$ represents the historical value of the observation and intervention time-series up to the prediction time $\x_{1:t^*}, \mathbf{u}_{1:t^*}$. Fig.~\ref{fig:compass} illustrates an example of our prediction task for four types of organ failure events and mortality with two two co-variants: non-invasive mean blood pressure (observation) and Dopamine drip rate (intervention). The prediction is made at time $8$ marked as vertical dashed lines, with forecast horizon up to time $30$. From the figure, we can see that the input features (i.e., co-variants) are represented as time-series data where values are measured at discrete time points\footnote{Note that the values from EMR are typically measured at irregular time intervals. We will explain in Sec.~\ref{sec:exp.preproc} how we handle such input data.}. The predicted event hazard rate vary over the forecast horizon with neural system failure hazard rate and coagulation failure hazard rate showing similar time-varying behaviors.
\section{Model Formulation}
\label{sec:model}
In this section, we first describe our proposed deep state-space model, augmented with intervention forecasting, which provides a principled way to capture the interactions among observations, interventions, patient states and their uncertainty. Based on this model, we further present a novel latent-state-generated hazard rate formulation for correlated time-to-event predictions. 

\begin{figure}[h]
\centering
        \begin{tikzpicture}[scale=0.8, transform shape]
        \node[latent] (z1) {$z_1$};
        \node[obs, below= of z1] (u1) {$u_1$};
        \node[obs,above= of z1] (x1) {$x_1$};
        \node[det,above right= of z1] (lmbd1) {$\lambda_1^e$};        
        \node[latent,right=of z1] (z2) {$z_2$};
        \node[obs, below=of z2] (u2) {$u_2$};
        \node[obs, above=of z2] (x2) {$x_2$};
        \node[det,above right= of z2] (lmbd2) {$\lambda_2^e$};        
        \node[latent,right=of z2] (z3) {$z_3$};
        \node[obs, below=of z3] (u3) {$u_3$};
        \node[obs, above=of z3] (x3) {$x_3$};
        \node[det,above right= of z3] (lmbd3) {$\lambda_3^e$};
        \node[latent,right=of z3] (z4) {$z_4$};
        \node[obs, above=of z4] (x4) {$x_4$};
        \node[det,above right= of z4] (lmbd4) {$\lambda_4^e$};
        \node[obs, below= of z4] (u4) {$u_4$};
        
	\edge{z1}{z2};
	\edge{z2}{z3};
	\edge{z3}{z4};
	\edge{z1}{x1};
	\edge{z2}{x2};
	\edge{z3}{x3};
	\edge{z4}{x4};
	\edge{z1}{lmbd1};
	\edge{z2}{lmbd2};
	\edge{z3}{lmbd3};
	\edge{z4}{lmbd4};
	%\edge{u1}{z2};
	\edge{u1}{z1};	
	\edge{u2}{z2};
	\edge{u3}{z3};
	\edge{u4}{z4};
	\edge{z1}{u2};
	\edge{z2}{u3};
	\edge{z3}{u4};
    \end{tikzpicture}
    \caption{\small Graphical Model of State-based Hazard Rate.}
\label{fig:gssm}
\end{figure}
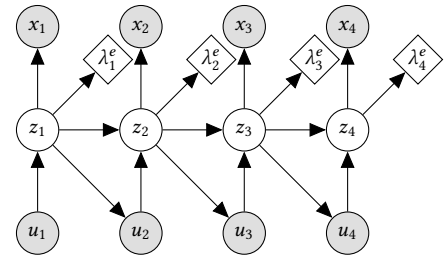

\begin{figure}[h]
\centering
        \begin{tikzpicture}[scale=0.8, transform shape]
        \node[latent] (z1) {$z_1$};
        \node[det,above right= of z1] (lmbd1) {$\lambda_1^e$};
        \node[det,above= of z1] (lmbd1p) {$\lambda_1^{e'}$};        
        \node[latent,right=of z1] (z2) {$z_2$};
        \node[det,above right= of z2] (lmbd2) {$\lambda_2^e$};
        \node[det,above= of z2] (lmbd2p) {$\lambda_2^{e'}$};        
        \node[latent,right=of z2] (z3) {$z_3$};
        \node[det,above right= of z3] (lmbd3) {$\lambda_3^e$};
        \node[det,above= of z3] (lmbd3p) {$\lambda_3^{e'}$};  
        \node[latent,right=of z3] (z4) {$z_4$};
        \node[det,above right= of z4] (lmbd4) {$\lambda_4^e$};
        \node[det,above= of z4] (lmbd4p) {$\lambda_4^{e'}$};
	\edge{z1}{z2};
	\edge{z2}{z3};
	\edge{z3}{z4};
	\edge{z1}{lmbd1};
	\edge{z2}{lmbd2};
	\edge{z3}{lmbd3};
	\edge{z4}{lmbd4};
	\edge{z1}{lmbd1p};
	\edge{z2}{lmbd2p};
	\edge{z3}{lmbd3p};
	\edge{z4}{lmbd4p};
    \end{tikzpicture}
    \caption{\small Graphical Model of Multi-Event-Type Hazard Rate.}
\label{fig:mevent}
\end{figure}
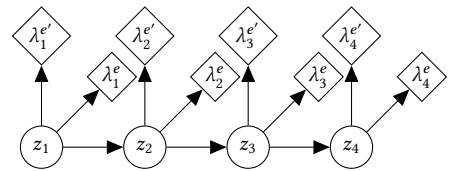

\subsection{State Space Model}
\label{sec:model.state}

To provide a joint time-to-event prediction of multiple clinical events, we need a powerful model that captures the temporal correlations among clinical observations and interventions. To this end, we adopt a Gaussian state space model to explicitly model the latent patient physiological state as shown in Fig.~\ref{fig:gssm}. Let $\mathbf{z}_t$ be the latent variable vector that represents the physiological state at time $t$ and $\mathbf{z}_{1:T}$ be the sequence of such latent variables. The system dynamics are defined via two equations:

\begin{align}
	&p(\mathbf{z}_t|\mathbf{z}_{t-1}, \mathbf{u}_{t}) \sim \mathcal{N} (\mathcal{A}_t(\mathbf{z}_{t-1}) + \mathcal{B}_t(\mathbf{u}_t), \bf{Q}) & \mathrm{Transition} \label{eqn:state_tran} \\ 
	&p(\mathbf{x}_t|\mathbf{z}_t) \sim \mathcal{N} (\mathcal{C}(\mathbf{z}_t), \bf{R}) &\mathrm{ Emission} \label{eqn:obs_emission}
\end{align}

Eq.~\eqref{eqn:state_tran} defines the state transition. Specifically, function $\mathcal{A}$ defines the system transition without external influence, i.e., how patient state will evolve from $\mathbf{z}_{t-1}$ to $\mathbf{z}_t$ without intervention. $\mathcal{B}$ captures the effect of intervention $\mathbf{u}_t$ on patient state $\mathbf{z}_t$. In Eq.~\eqref{eqn:obs_emission}, $\mathcal{C}$ captures the relationship between internal state $\mathbf{z}_t$ and observable measurements $\mathbf{x}_t$. $\mathbf{Q}$ and $\mathbf{R}$ are process and measurement noise covariance matrices. We assume them to be time-invariant. Eq.~\eqref{eqn:state_tran} and \eqref{eqn:obs_emission} subsume a large family of linear and non-linear state space models. For example, by setting $\mathcal{A}, \mathcal{B}, \mathcal{C}$ to be matrices, we obtain linear state space models. By parameterizing $\mathcal{A}, \mathcal{B}, \mathcal{C}$  via deep neural networks, we have a deep state space model. 

\textit{Intervention Forecast}. Contrary to classical state space models, where interventions are usually considered as external factors, when inferring patient states from EMR data, interventions are an integral part of the system, as they are determined by clinicians based on their estimation of patient states and medical knowledge/clinical guidelines. To model this relationship, we augment the state space model with additional dependency from $\mathbf{z}_t$ to $\mathbf{u}_{t+1}$ as shown in Fig.~\ref{fig:gssm}.

\begin{equation}
p(\mathbf{u}_t|\mathbf{z}_{t-1}) \sim \mathcal{N}(\mathcal{D}(\mathbf{z}_{t-1}), \bf{U})  \label{eqn:int_emission}
\end{equation}

Similarly, in Eq.(\ref{eqn:int_emission}) $\mathcal{D}$ can be either a matrix for a linear model or parameterized by a neural network for a nonlinear model. For clinical predictions, there are two different questions one may ask: 1) what will happen if \textit{no} intervention is applied; 2) what will happen if the patient receives expected interventions. Our model allows us to answer the second question. 

\subsection{State-based Discrete-time Hazard Rate}
\label{sec:model.survival}

Recall that the hazard rate function describes the instantaneous rate of event occurrence at time $t$. In classical survival analysis, this rate is usually assumed to be constant over time and statically determined by the co-variants at the time of prediction~\cite{cox}. Based on the physiological state-space model, we propose a new time-to-event estimation model where the hazard rate function is discretized per time step and dependent on the dynamic latent patient physiological state at that time. Specifically, the hazard rate $\lambda^e_{t}$ of event $e$ at time step $t$ is modelled as
\begin{equation}
\lambda_{t}^e = \mathcal{L}^e(\mathbf{z}_t) 
\end{equation}
where  $\mathcal{L}^e$ can be either a linear model or neural network to map the hidden variable $\mathbf{z}_t$ to a deterministic value. The discrete survival function at time $t$ can be written as
\begin{equation}
    S^e(t)  = (1 - \lambda^e_t) S^e(t-1).
\end{equation} Let $S^e(0) = 1$. The above recursion leads to 
\begin{equation}
S^e(t) = \prod_{s = 1}^{t} (1- \lambda^e_s).
\end{equation}
The incidence density function is defined as $f(t^e) = \Pr(t^e = t)$ and is connected with $\lambda^e_t$ via
\begin{equation}
f(t^e) = \lambda^e_t \prod_{s = 1}^{t-1} (1- \lambda^e_s).
\end{equation}

All event $e\in E$ are generated from the shared states $\mathbf{z}_t$ but with its individual generation function $\mathcal{L}^e$. Fig.~\ref{fig:mevent} shows a graph model for two events $e, e'$. Note that observation and intervention nodes are omitted in this figure for clear illustration.
%The expected survival time can be obtained as \[\mathbb{E}(T) = \sum_t \Pr(T > t) = \sum_t S(t). \]

\section{Variational Inference}
Our state space model is fully specified by the generative parameter $\theta = (\mathcal{A}, \mathcal{B}, \mathcal{C}, \mathcal{D}, \mathcal{L}^e, e\in E)$. In this section, we present the learning objective and the associated variational lower bound that supports the time-to-event prediction task as described in Sec.~\ref{sec:problem}.

Recall that time to event prediction estimates the time distribution of $t^e$ at $t^*$ based on the historical values of $\bar{\x}, \bar{\mathbf{u}}$. We first consider the log likelihood of one event $e$ represented by $(c,t^e)$. There are two cases for this event: 
\begin{itemize}
    \item if $c=1$, which means the event is censored/survived at $t^e$, then the likelihood is captured by its survival function $\log S_\theta(t^e |  \bar{\x}, \bar{\mathbf{u}})$;
    \item if $c=0$, which means the event is observed at $t^e$, then the likelihood is captured by its incidence density function $f_\theta(t^e |  \bar{\x}, \bar{\mathbf{u}})$.
\end{itemize} 

Further recall that $\lambda^e_{t} = \mathcal{L}^e(\z_t)$ and $S^e(t^e) = \prod_{s = 1}^{t^e} [1- \lambda^e_s]$, $f^e(t^e) = \prod_{s = 1}^{t^e-1} [1- \lambda^e_s]\cdot \lambda^e_{t^e}$. Thus $f^e_\theta(t^e)$ and $\log S^e_\theta(t^e)$ are independent from observations and interventions conditioned on hidden state $\hat{\z}$, where $\hat{\z} = \z_{1:t^e}$. Putting both cases together and based on the graph model in Fig.~\ref{fig:gssm}, we have the log likelihood of $e$ as follows:

\begin{eqnarray*}
\label{eqn:survival_likelihood}
\log \pth(t^e | \bar{\x}, \bar{\mathbf{u}}) &=& 
\underbrace{(1-c) \cdot \log f^e_\theta(t^e |  \bar{\x}, \bar{\mathbf{u}})}_\text{event is observed at $t^e$} \\
&+& 
\underbrace{c \cdot \log S^e_\theta(t^e |  \bar{\x}, \bar{\mathbf{u}}) }_\text{$e$ is censored/survived at $t^e$} \\
&=& (1-c) \cdot \log \int_{\hat{\z}} \pth(\hat{\z} | \bar{\x}, \bar{\mathbf{u}}) f^e_\theta(t^e|\hat{\z}) \\
&+& c \cdot \log \int_{\hat{\z}} \pth(\hat{\z} | \bar{\x}, \bar{\mathbf{u}}) S^e_\theta(t^e|\hat{\z})
\end{eqnarray*}

This log likelihood is intractable when inferring the posterior $\pth(\hat{\z}|\bar{\x}, \bar{\mathbf{u}})$. We adopt the variational inference method by introducing a variational distribution $\qph$ that approximates this posterior. The evidence lower bound (ELBO) of the log event time likelihood is thus given as:

\begin{align}
\label{eq:survival_loss}
& \underbrace{(1-c) \cdot \Exp{\log f^e_\theta(t^e|\hat{\z})}{\qph(\hat{\z}|\bar{\x}, \bar{\mathbf{u}})} + c \cdot \Exp{\log  S^e_\theta(t^e|\hat{\z})}{\qph(\hat{\z}|\bar{\x}, \bar{\mathbf{u}})}}_\text{time-to-event prediction loss} \\
& \underbrace{-\mathbb{KL}(\, \qph(\hat{\z}| \bar{\x},\bar{\mathbf{u}}) || \pth(\hat{\z} | \bar{\x}, \bar{\mathbf{u}}))}_\text{regularization loss} \nonumber
\end{align}

The lower bound in Eq.(\ref{eq:survival_loss}) has two components: 1) the log likelihood loss for time-to-event prediction; 2) the regularization loss which measures the difference between the encoder and the simple prior distribution of the latent state $\z$ given the transition model between $z_{t-1}$ and $z_t$ as defined in the state space model (Eq.(\ref{eqn:state_tran})). Similar to~\cite{nipsw15-dkf}, this ELBO can be factorized along time as:

\begin{align}
\label{eq:elbo}
&(1-c) \cdot \Exp{\sum_{s=1}^{t^e-1}\log (1-\mathcal{L}^e(\z_t))  +\mathcal{L}^e(\z_t)}{\qph(\z_t|\bar{\x}, \bar{\mathbf{u}})} \nonumber\\
&+ c \cdot \Exp{\sum_{s=1}^{t^e}\log (1-\mathcal{L}^e(\z_t))}{\qph(\z_t|\bar{\x}, \bar{\mathbf{u}})} \nonumber\\
&  - \sum_{t=1}^{t^e} \mathbb{KL}(\, \qph(\z_t| \z_{t-1}, \bar{\x}, \bar{\mathbf{u}}) || \pth(\z_t| \z_{t-1}, \bar{\mathbf{u}})\, ) 
\end{align}

For a set of events $e$, the loss function is a sum of all the negative log event time likelihood: $-\sum_{e\in E} \log \pth(t^e | \bar{\x}, \bar{\mathbf{u}})$, each of which is based on the same latent state estimation and the hazard rate generation function associated with its event type.

\section{Model Architecture}

\begin{figure*}[!hbtp]
\begin{center}
\includegraphics[scale=0.30]{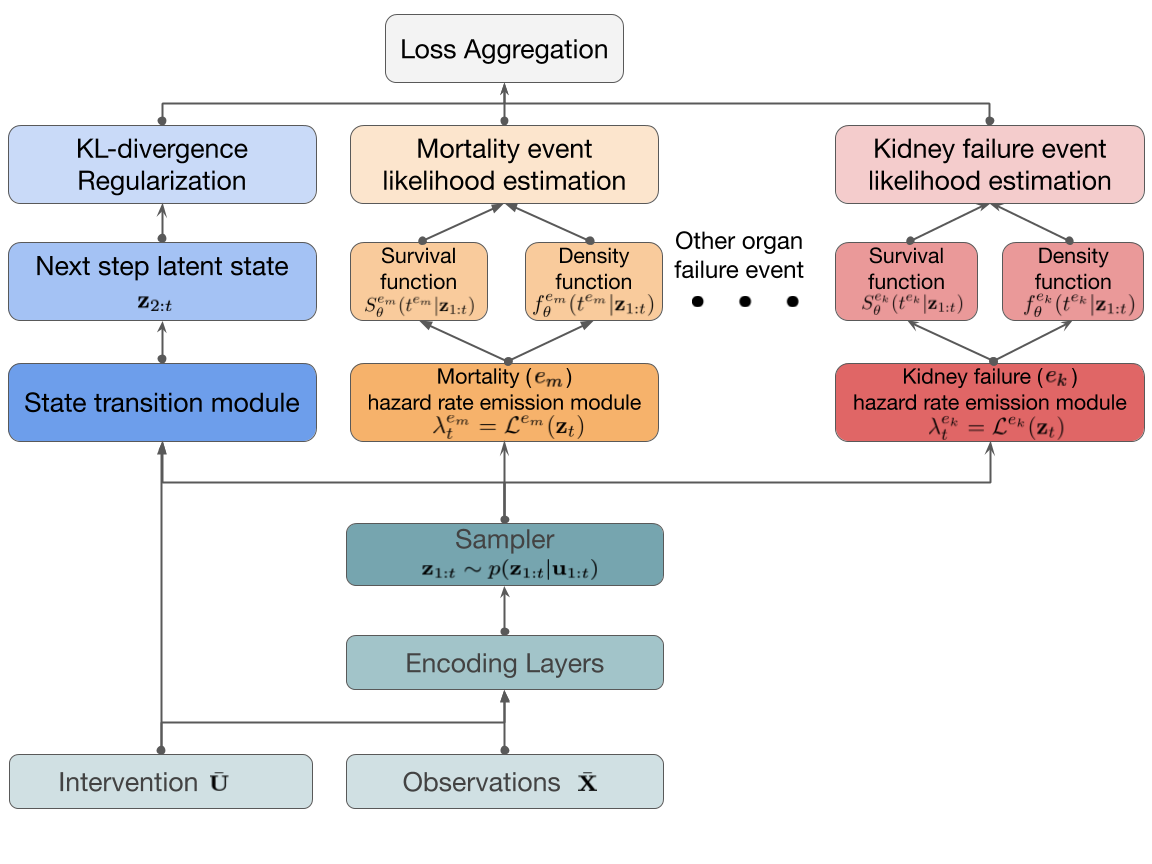}
\end{center}
\caption{Loss Computation Process. The state-space computation task (leftmost) captures the changing dynamics of patients' latent states $\mathbf{z}_{1:t}$ based on past observations and interventions. Each following task of event prediction (mortality, kidney failure, etc.) has it own survival and density function that depend on these latent states $\mathbf{z}_{1:t}$.}
\label{fig:model_arch}
\end{figure*}

We describe our learning algorithm and the neural network models used for learning in this section. As shown in Fig.~\ref{fig:model_arch}, give the ELBO, our learning algorithm proceeds the following steps: 
\begin{itemize}
    \item Inference of $\hat{\mathbf{z}}$ from $\bar{\mathbf{x}}$ and $\bar{\mathbf{{u}}}$ by an encoder network $\qph$. We follow the same model architecture as in ~\cite{nipsw15-dkf} and use a bi-directional LSTM as the encoder network.
    \item Sampling based on the current estimate of the posterior $\hat{\mathbf{z}}$.
    \item Estimate the next step latent state $\z_{2:t}$ via the generative model $\pth$ and compute the regularization loss. We use two multi-layer perceptrons (MLP) for the state transition module -- one for  state transition without external influence; the other for the effect of intervention on state transition. 
    \item Estimate the hazard rate for each type of event of interest via the generative model $\pth$. Each event has a separate hazard rate emission module which is a MLP. 
    \item Survival function and incidence density function are computed based on the estimated hazard rate, from which the negative log event likelihood loss is computed for each types of event.
    \item The likelihood loss of all events and the KL-divergence regularization loss are aggregated as the training loss (negative ELBO). 
    \item Estimate the gradients of the loss with respect to $\theta$ and $\phi$ and updating parameters of the model. Gradients are averaged stochastically across  mini-batches of the training set.  
\end{itemize} 

Our model is implemented in TensorFlow~\cite{tensorflow2015-whitepaper}, and will be open-sourced~\footnote{https://github.com/Google-Health/records-research/state-space-model}.

\section{Experiments}
We extensively evaluate our proposed deep state-space model (DSSM) over real temporal event data showing that it has better predictive performance for time-to-event prediction, and is able to uncover meaningful insights about the latent correlation among different types of events. 
\begin{figure*}%
    \centering
    {
    {\includegraphics[width=18cm]{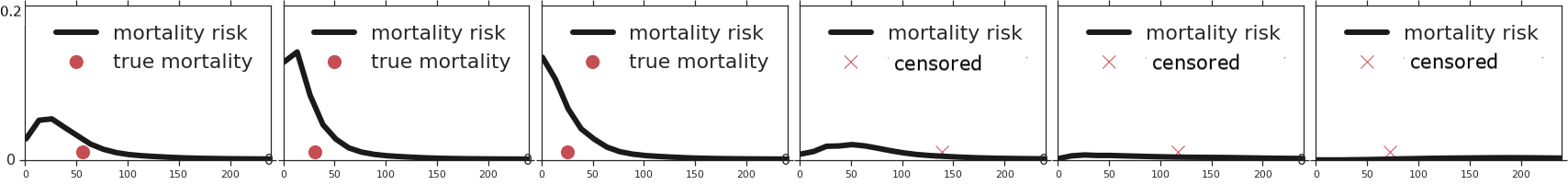} }}%
    \qquad
    {
    {\includegraphics[width=18cm]{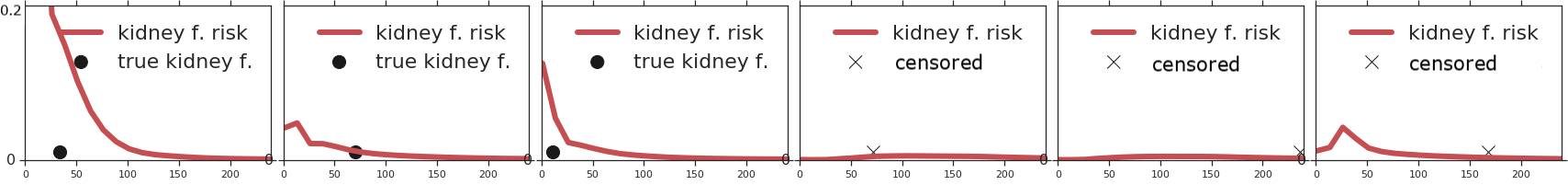} }}%
    \qquad
    {
    {\includegraphics[width=18cm]{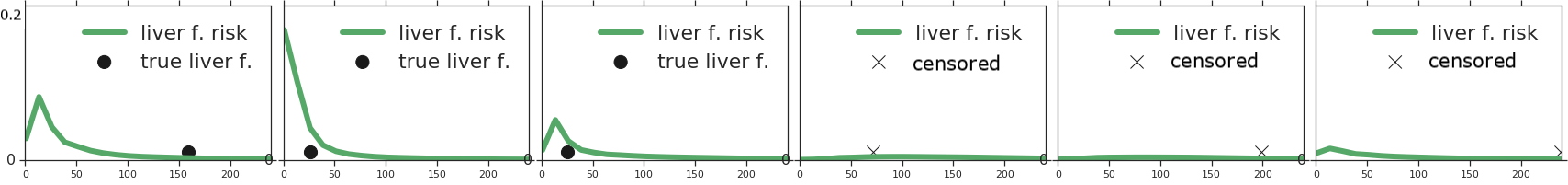} }}%
    \qquad
    {
    {\includegraphics[width=18cm]{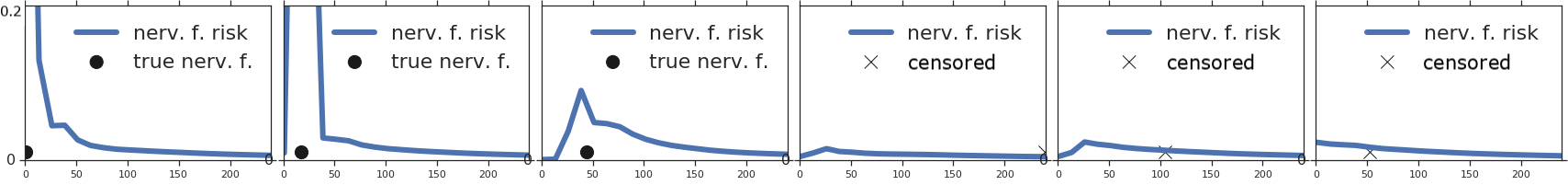} }}%
    \qquad
    {
    {\includegraphics[width=18cm]{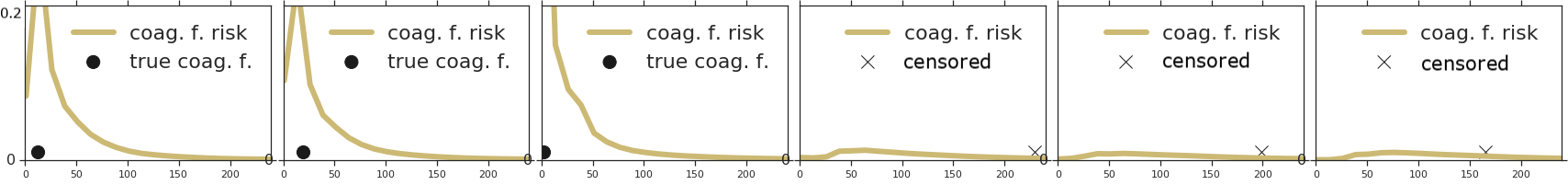} }}%
%    \qquad
%    {
%    {\includegraphics[width=18cm]{figs/nerv.png} }}%
    \caption{Event hazard rate and true occurrence time for both observed (in solid dot) and censored (in light cross) events. For each prediction task, the fitted hazard rate is able to capture the true event time accurately, while for the censored events, the respective hazard rates are low as expected.}%
    \label{fig:single_traj}%
\end{figure*}

\begin{figure*}[h!]
  \includegraphics[scale=0.33]{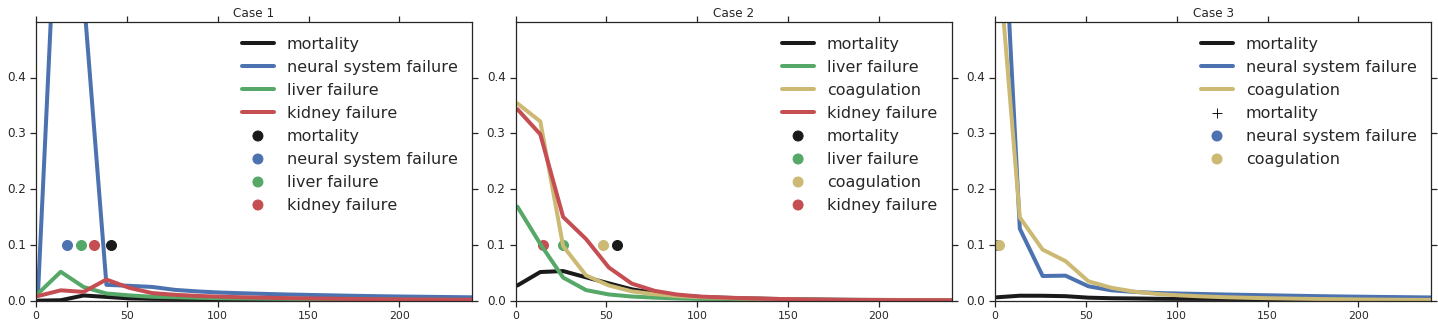}
  \caption{Different correlated trajectories of Multi-Organ Failure and Mortality. In Case 1 and 2, mortality is highly correlated with other organ failures as indicated by the learned hazard rates. In contrast, for Case 3, the neural system failure and coagulation have little influence on mortality, which is also reflected in the learned hazard rate curve.}
  \label{fig:mof}
\end{figure*}

\vspace{-1mm}
\subsection{Dataset and Data Preprocessing}
\label{sec:exp.preproc}

\begin{table}[t]
\centering
\small
\begin{tabular}{l c c c} \\
\toprule 
                & \#Positive & \#Negative &\#Excluded \\
 \midrule
Mortality &  4,277 & 41,843 & 4,557  \\ 
\midrule
Kidney Failure &  2,056 &  33,655 & 14,966 \\ 
\midrule
Liver Failure &  1,474 & 39,469 &  9,734  \\ 
\midrule
Coagulation Failure &  2,496 &  35,496 &  12,685 \\ 
\midrule
Nervous system Failure &  3,475 &  28,340 &  18,862\\ 
 \bottomrule
\end{tabular}
\vspace{3mm}
\caption{Statistics of different event prediction tasks.}
\vspace{-8mm}
\label{table:label_stat}
\end{table}

We use Medical Information Mart for Intensive Care (MIMIC) data~\cite{johnson16mimic} in our empirical study. MIMIC-III is a large open EMR dataset containing information relating to patients admitted to critical care units at a large tertiary care hospital. Data includes vital signs, medications, laboratory measurements,  procedure codes, diagnostic codes, and more.

We consider inpatients from MIMIC-III who are still alive $48$ hours after admission and predict their risk of in-hospital death at $48$ hours after admission along with the risk of $4$ multiple organ failures based on the SOFA score~\cite{sofa_score} definition:
\begin{itemize}
    \item \textbf{Kidney Failure}: creatinine $\geq2$mg/dl
    \item \textbf{Liver Failure}: bilirubin $\geq2$mg/dl
    \item \textbf{Coagulation Failure}: platelet $< 100 \times 10^3$/ $\mu l$
    \item \textbf{Nervous system Failure}: Glasgow coma scale $\leq 12$
\end{itemize}    

As organ failures can be recurrent, we only consider the first occurrence of each type of organ failure in an encounter as the event of interest. The statistics of these events from MIMIC-III are provided in Table~\ref{table:label_stat}. The positive (negative) columns are the number of encounters where the corresponding events are observed (unobserved) after the prediction time. The excluded column counts the encounters where the corresponding event occurs before the prediction time. We only include the encounters free of any organ failures in our study. There are $38485$ adult in-patient encounters included in the study %with $3145$ observed in hospital death.
The mean length of stay of the encounters is around $10$ days. As a side note, the high number of excluded encounters is related to the strong association between organ failures and ICU admission. And the organ failures we included in our study are usually newly developed after ICU admission.

We select $96$ most frequently used lab measurements and vital signs as observation features, $8$ types of vasopressors and antibiotics and $6$ most recorded ventilation and dialysis machine settings as intervention features. For intervention features, we consider the following parameters to indicate the presence and the level of medical interventions: 1) the dosage and drip rate of medication administrations, 2) mechanical ventilation and dialysis machine settings. 

As different coding systems are used in MIMIC-III, we harmonize the medical codes corresponding to the same lab/vital measurement as a single feature. In addition, we standardize the unit when a medical code is used with multiple units or without a unit. All observation and intervention values are normalized using z-score, where the mean and standard deviation of each feature are computed based on the MIMIC-III dataset. The outlier measurements are removed from training. The details of feature selection and data prepossessing are reported in the supplemental material.

Observational data is recorded at irregular intervals in EMR, resulting in a large number of missing values when sampled at regular time steps. Handling missing value in observation data has been investigated in recent works~\cite{Che2018-vj}. For lab measurements and vital signs, we adopt a simple method where the most recent value is used to impute the missing ones. For interventions, the situation is more complex and not handled in any existing works. Specifically, we need to differentiate the case where a missing value represents that the intervention is not performed or has been completed vs. the case where a missing value means the same setting is continued at this time step. To address this issue, we follow the observation that most (continuous) medication are administrated at regular intervals and organ support machine settings are also regularly adjusted. We first derive the distribution of inter-medication-administration and inter-intervention-setting time, then pick the $90$-percentile time as the cut-off threshold. If two consecutive interventions are within the time range of their corresponding thresholds, then we consider the missing value as an indication of a continuous action and use the last setting as its missing value. If it falls outside of this range, then a missing value is considered as no action.

\vspace{-2mm}
\subsection{Prediction Performance}

We first evaluate the overall performance of time-to-event predictions for each event. The following metrics are used: 

\begin{itemize}
\item \textbf{C-index} (i.e., concordance index) measures the extent to which the ordering of actual event times of pairs agrees with the ordering of their predicted risk. It is a widely used discriminative metric for evaluating the performance of survival models. 
\item \textbf{AUC-ROC and AP} (a.k.a. AUPRC) within two fixed prediction windows [$0, 24$]hr and [$0, 48$]hr. This metric evaluates the short-term prediction performance, while C-index evaluates the overall model prediction power. To accurately compute AUC-ROC and AP for an event prediction within a fixed time window in the presence of censorship, we only consider 1) the events which are observed within this window as positives, and 2) the events which are either observed or censored outside of the window as negatives. These are the cases where we can be sure that the event does not occur in the window. If an event is censored within the window (e.g., the patient is discharged within the window without the specified event being observed), it is not included in the computation.
\end{itemize}

We compare our methods with two state-of-art methods:

\begin{itemize}
\item \textbf{DeepSurv}~\cite{bmc18-katzman-deepsurv} is a Cox proportional hazards deep neural network that
models the interactions between a patient’s covariates and the outcome. DeepSurv model uses a MLP to first encode the co-variants then use them as the coefficients in the Cox model. In our experiment, we use a 3-layer MLP with sigmoid activation as the encoder and the most recent observation values of each feature as the co-variants.

\item \textbf{DRSA}~\cite{aaai19-ren} refers to Deep Recurrent Survival Analysis, which is a model based on the recurrent neural network. In this model, the $l$-th RNN cell predicts the instantaneous hazard rate at time step $l$. This method can be considered as the deterministic counterpart of our deep state space encoding. The key difference between our work is that there is no regularization loss as in Eq(\ref{eq:survival_loss}) in DRSA. In addition, DRSA considers a static co-variant $\mathbf{x}$, while our model considers the time series of observations and interventions as features from which the intrinsic dynamics of the hidden states are learned and used to forecast the hazard rate. For fair comparison, in our experiment, we use an improved version of DRSA, where the time series of observation and intervention values are encoded using LSTM as co-variant $\mathbf{x}$.

\item \textbf{DSSM} is our proposed deep state space model. This model encodes the temporal correlations among observations, interventions and hidden states, which are rolled-out to the prediction horizon step by step to generate the hazard rate for each event.

\end{itemize}

The hyperparameters including the learning rate, the hidden state size for LSTM, the number of units and layers for MLP, the size of time step are tuned. The experiment uses a hidden state size of $50$ for LSTM and $32$ hidden units with $3$ layers for all MLPs, including the state transition MLP, the intervention effect MLP and observation emission MLP. A learning rate of $0.0001$ is used for \textbf{DSSM} and \textbf{DRSA} and learning rate of $0.0005$ is used for \textbf{DeepSurv}. In the experiment, each time step takes $12$ hours. The prediction rolls out to $240$ time steps, beyond which a constant hazard rate which is the same as the last time step is used.

To demonstrate the performance variance in our evaluation, we use $10$-fold cross validation. For each fold, we split the dataset into train/eval/test according to $80\%$/$10\%$/$10\%$ based on the hash value of the patient ID. We estimate the standard error of the mean based on the sample standard deviation of these $10$ folds. Table~\ref{table:survival_result} reports the mean and the standard error for the three models. We can see that our proposed method \textbf{DSSM} outperforms \textbf{DeepSurv} and \textbf{DRSA} on all the metrics. In addition, our model brings significant improvement on the short-term predictions in terms of both AUC-ROC and AP, as the state roll-outs tend to be more accurate at the closer forecast horizon. Though \textbf{DRSA} also rolls out the state of a LSTM for hazard rate prediction, its performance is limited by the LSTM's tendency to use recent history. While our approach incorporates the regularization loss that minimizes the KL divergence between the encoded state and the prior distribution of the latent state under the state transition model, which encourages the true state dynamics to be learnt.

\vspace{-2mm}
\subsection{Hazard Rate Trajectory}

We now zoom in to individual patients and study the hazard rate dynamics. Fig.~\ref{fig:single_traj} plots the hazard rate trajectory within the first $240$hr after the prediction time for mortality and organ failures separately on each row. All plots share the same y-axis range $[0, 0.2]$. The first three figures on each row plot the cases where the corresponding event happens to the patient. The true event time is plotted in the figure as a dot. The last three figures plot the cases where the event is not observed and the censor time (when the patient is discharged) is plotted as a cross. 

Comparing these figures, we can clearly see that the hazard rate is significantly higher for the observed cases than the cases where events are not happened in the observation window. In addition, the flexible discretized-time model provides more accurate instantaneous hazard rate estimation, as reflected in the fact that a large portion of the events happen around the time where the  hazard rate is peaked. This is in contrast with the constant hazard assumption which is usually made in the conventional survival analysis\cite{cox}. Lastly we observe that all the trajectories tend to converge to a base value after around $150$ hrs due to the fact that no new data comes in after the prediction point. This baseline values vary for different patients and different events.

\vspace{-2mm}
\subsection{Hazard Rate Correlations}

\begin{table}[t]
\centering
\small
\begin{tabular}{l c c c c} \\
\toprule 
                & Neural. & Liver & Kidney & Coagulation\\
 \midrule
Case 1 &  0.596 & 0.745 & 0.741 & 0.577 \\ 
\midrule
Case 2 &  0.810 &  0.838 & 0.914 & 0.854 \\ 
\midrule
Case 3 &  0.477 &  0.565 &  0.784 & 0.5270\\ 
 \bottomrule
\end{tabular}
\vspace{3mm}
\caption{Correlations between Mortality and Organ Failures.}
\vspace{-8mm}
\label{table:corr}
\end{table}

We further show that our model can reveal the correlations among these events which can in turn facilitate the understanding of the progression of patient conditions. Fig.~\ref{fig:mof} plots the hazard rate trajectories along with their true event time of three patient encounters -- Case 1 and 2 corresponding to mortality cases and Case 3 corresponding to a survived case. For presentation clarity, only the organ failures that happened within the window of $240$hr are plotted. For both mortality case, multiple organ failures are predicted with high hazard rate around the prediction of the increased mortality hazard rate. This prediction is validated by the sequence of true organ failure and mortality. For Case 3, the neural system failure and coagulation failure are predicted at high risk at the beginning of the prediction horizon. The true event time for both failures are $1$hr after prediction time. They increase the mortality risk slightly at the beginning. All three risks -- neural system failure, coagulation failure and mortality drop significantly and the patient is discharged at $266$hr.

From the figure, we can also see that the mortality hazard rate trajectory is strongly correlated with the trajectory of organ failures. In order to quantify the correlation, we cross-correlate organ failure trajectories with mortality trajectory and show the correlation coefficients in Table.~\ref{table:corr}. Though mortality has positive correlations with all organ failures, the level of correlation varies by individual patient and different organs, which reveals valuable insights. In Case 1, mortality is predicted to be highly correlated with the kidney and liver failure, and both are also predicted with high predicted risks. Similarly, in Case 2, mortality is highly correlated with liver, kidney, coagulation failures, and all are predicted with high risks. This provides an explanation for the mortality events in addition to the mortality risk prediction. In contrast, for Case 3, though the neural system failure and coagulation failure are predicted to have high risk, the correlation between mortality and neural system and coagulation is relatively low as shown in Table.~\ref{table:corr}, indicating mortality is less influenced by these two organ failures in this case. 

Comparing with the single mortality event prediction, our correlated predictions provide insights into why mortality may happen. This offers clinicians with a full picture of a patient’s medical condition and better supports them with better decision making.

\begin{table}[t]
%\footnotesize
\scriptsize
%\resizebox{\columnwidth}{!}{
\centering
\begin{tabular}{l p{0.8cm}   p{0.8cm} p{0.8cm}   p{0.8cm}   p{0.8cm}}  \\
%\begin{tabular}{c  }  \\
\toprule 
                & C-index& AUC@24 & AP@24 & AUC@48 &  AP@48 \\
 \midrule
 Mortality  & & & & &  \\
\cmidrule{1-6}
DeepSurv &	0.769 (0.009)&	0.911 (0.015)&	0.25 (0.056)&	0.861 (0.014)&	0.228 (0.043)\\
DRSA  &	0.743 (0.016)&	0.906 (0.025)&	0.28 (0.0092)&	0.837 (0.015)&	0.186 (0.031)\\
DSSM &	\textbf{0.7769} (0.007)&	\textbf{0.949} (0.021)&	\textbf{0.375} (0.091)&	\textbf{0.873} (0.016)&	\textbf{0.258} (0.036) \\
\cmidrule{1-6}
Kidney Failure  & & & & &  \\
\cmidrule{1-6}
DeepSurv &	0.826 (0.007)&	0.957 (0.007)&	0.315 (0.037)&	0.901 (0.009)&	0.319 (0.033) \\
DRSA  &	0.810 (0.007)&	0.944 (0.011)&	0.221 (0.039)&	0.876 (0.011)&	0.221 (0.035)\\
DSSM &	\textbf{0.829} (0.003)&	\textbf{0.981} (0.005)&	\textbf{0.485} (0.057)&	\textbf{0.914} (0.005)&	\textbf{0.365} (0.033)\\
\cmidrule{1-6}
Liver Failure  & & & & &  \\
\cmidrule{1-6}
DeepSurv &	\textbf{0.709} (0.015)&	0.752 (0.018)&	0.062 (0.017)&	0.733 (0.015)&	0.071 (0.020) \\
DRSA  &	0.702 (0.017)&	0.778 (0.025)&	0.056 (0.021)&	0.745 (0.026)&	0.056 (0.017)\\
DSSM &	\textbf{0.709} (0.012)&	\textbf{0.843} (0.019)&	\textbf{0.132} (0.032)&	\textbf{0.784} (0.019)&	\textbf{0.101} (0.002)\\
\cmidrule{1-6}
Coagulation Failure  & & & & &  \\
\cmidrule{1-6}
DeepSurv &	0.831 (0.012)&	0.875 (0.020)&	0.239 (0.041)&	0.863 (0.019)&	0.265 (0.031)\\
DRSA  &	0.803 (0.007)&	0.928 (0.011)&	0.196 (0.029)&	0.861 (0.009)&	0.216 (0.015) \\
DSSM &	\textbf{0.835} (0.007)&	\textbf{0.942} (0.007)&	\textbf{0.292} (0.039)&	\textbf{0.890} (0.009)&	\textbf{0.272} (0.019) \\
\cmidrule{1-6}
Neural Sys. Failure  & & & & &  \\
\cmidrule{1-6}
DeepSurv &	0.852 (0.003)&	0.907 (0.005)&	0.587 (0.013)&	0.876 (0.004)&	0.552 (0.006) \\
DRSA  &	0.849 (0.003)&	0.948 (0.01)&	0.68 (0.026)&	0.870 (0.007)&	0.526 (0.012)\\
DSSM &	\textbf{0.863} (0.004)&	\textbf{0.968} (0.006)&	\textbf{0.751} (0.02)&	\textbf{0.889} (0.005)&	\textbf{0.586} (0.008)\\
 \bottomrule
\end{tabular}
%}
\vspace{3mm}
\caption{\small Time-to-Mortality Prediction Performance. Parentheses denote standard error.}
\vspace{-8mm}
\label{table:survival_result}
\end{table}

\section{Conclusions}

We proposed a deep latent state-space generative model to capture the relations between patients' mortality risk and the associated organ failure risks. Based on the learned patients' states, we further develop a new formulation of the hazard rate function to fit general discrete-time survival distribution of observed events. Extensive experiments over MIMIC datasets show that our proposed model not only outperforms several state-of-art baselines in terms of prediction accuracy, but also provides meaningful insights into the temporal relations among the multiple types of events. By demonstrating the correlations between different organ failures and mortality risk, we provide physicians with more evidence to have better decision-making.

\bibliographystyle{ACM-Reference-Format}
\bibliography{ref}


\begin{thebibliography}{36}

%%% the Web address.

\ifx \showCODEN    \undefined \def \showCODEN     #1{\unskip}     \fi
\ifx \showDOI      \undefined \def \showDOI       #1{#1}\fi
\ifx \showISBNx    \undefined \def \showISBNx     #1{\unskip}     \fi
\ifx \showISBNxiii \undefined \def \showISBNxiii  #1{\unskip}     \fi
\ifx \showISSN     \undefined \def \showISSN      #1{\unskip}     \fi
\ifx \showLCCN     \undefined \def \showLCCN      #1{\unskip}     \fi
\ifx \shownote     \undefined \def \shownote      #1{#1}          \fi
\ifx \showarticletitle \undefined \def \showarticletitle #1{#1}   \fi
\ifx \showURL      \undefined \def \showURL       {\relax}        \fi
% invisible to TeX
\providecommand\bibfield[2]{#2}
\providecommand\bibinfo[2]{#2}
\providecommand\natexlab[1]{#1}
\providecommand\showeprint[2][]{arXiv:#2}

\bibitem[Abadi and et~al.(2015)]%
        {tensorflow2015-whitepaper}
\bibfield{author}{\bibinfo{person}{Mart\'{\i}n Abadi} {and} \bibinfo{person}{et
  al.}} \bibinfo{year}{2015}\natexlab{}.
\newblock \bibinfo{title}{TensorFlow: A System for Large-Scale Machine
  Learning}.
\newblock
\newblock


\bibitem[Alaa and van~der Schaar(2017)]%
        {nips17-schaar-gaussian}
\bibfield{author}{\bibinfo{person}{Ahmed~M. Alaa} {and}
  \bibinfo{person}{Mihaela van~der Schaar}.} \bibinfo{year}{2017}\natexlab{}.
\newblock \showarticletitle{Deep Multi-task Gaussian Processes for Survival
  Analysis with Competing Risks}.
\newblock In \bibinfo{booktitle}{\emph{NIPS}}. \bibinfo{pages}{2329--2337}.
\newblock


\bibitem[Alaa and van~der Schaar(2019)]%
        {nips19-schaar-attentive}
\bibfield{author}{\bibinfo{person}{Ahmed~M. Alaa} {and}
  \bibinfo{person}{Mihaela van~der Schaar}.} \bibinfo{year}{2019}\natexlab{}.
\newblock \showarticletitle{Attentive State-Space Modeling of Disease
  Progression}. In \bibinfo{booktitle}{\emph{NeurIPS}}.
\newblock


\bibitem[Barretta and Coolena(2010)]%
        {sim10-barrett-gaussian}
\bibfield{author}{\bibinfo{person}{James~E. Barretta} {and}
  \bibinfo{person}{Anthony C.~C. Coolena}.} \bibinfo{year}{2010}\natexlab{}.
\newblock \showarticletitle{Gaussian process regression for survival data with
  competing risks}.
\newblock


\bibitem[Bellot and van~der Schaar(2018)]%
        {nips18-schaar-boosting}
\bibfield{author}{\bibinfo{person}{Alexis Bellot} {and}
  \bibinfo{person}{Mihaela van~der Schaar}.} \bibinfo{year}{2018}\natexlab{}.
\newblock \showarticletitle{Multitask Boosting for Survival Analysis with
  Competing Risks}.
\newblock In \bibinfo{booktitle}{\emph{NeurIPS}}.
\newblock


\bibitem[Chapfuwa et~al\mbox{.}(2018)]%
        {icml-date}
\bibfield{author}{\bibinfo{person}{Paidamoyo Chapfuwa},
  \bibinfo{person}{Chenyang Tao}, \bibinfo{person}{Chunyuan Li},
  \bibinfo{person}{Courtney Page}, \bibinfo{person}{Benjamin Goldstein},
  \bibinfo{person}{Lawrence Carin}, {and} \bibinfo{person}{Ricardo Henao}.}
  \bibinfo{year}{2018}\natexlab{}.
\newblock \showarticletitle{Adversarial Time-to-Event Modeling}. In
  \bibinfo{booktitle}{\emph{ICML}}.
\newblock


\bibitem[Che et~al\mbox{.}(2018)]%
        {Che2018-vj}
\bibfield{author}{\bibinfo{person}{Zhengping Che}, \bibinfo{person}{Sanjay
  Purushotham}, \bibinfo{person}{Kyunghyun Cho}, \bibinfo{person}{David
  Sontag}, {and} \bibinfo{person}{Yan Liu}.} \bibinfo{year}{2018}\natexlab{}.
\newblock \showarticletitle{Recurrent Neural Networks for Multivariate Time
  Series with Missing Values}.
\newblock \bibinfo{journal}{\emph{Sci. Rep.}} \bibinfo{volume}{8},
  \bibinfo{number}{1} (\bibinfo{year}{2018}).
\newblock


\bibitem[Choi et~al\mbox{.}(2015)]%
        {Choi2015-ak}
\bibfield{author}{\bibinfo{person}{Edward Choi}, \bibinfo{person}{Mohammad~Taha
  Bahadori}, \bibinfo{person}{Andy Schuetz}, \bibinfo{person}{Walter~F
  Stewart}, {and} \bibinfo{person}{Jimeng Sun}.}
  \bibinfo{year}{2015}\natexlab{}.
\newblock \showarticletitle{Doctor {AI}: Predicting Clinical Events via
  Recurrent Neural Networks}.
\newblock  (\bibinfo{date}{Nov.} \bibinfo{year}{2015}).
\newblock
\showeprint{1511.05942}


\bibitem[Cox(1992)]%
        {cox}
\bibfield{author}{\bibinfo{person}{D.~R. Cox}.}
  \bibinfo{year}{1992}\natexlab{}.
\newblock \showarticletitle{Regression models and life-tables}.
\newblock In \bibinfo{booktitle}{\emph{Breakthroughs in statistics}}.
  \bibinfo{pages}{527–541}.
\newblock


\bibitem[Fernandez et~al\mbox{.}(2016)]%
        {nips16-fernandez-gaussian}
\bibfield{author}{\bibinfo{person}{Tamara Fernandez}, \bibinfo{person}{Nicolas
  Rivera}, {and} \bibinfo{person}{Yee~Whye Teh}.}
  \bibinfo{year}{2016}\natexlab{}.
\newblock \showarticletitle{Gaussian Processes for Survival Analysis}.
\newblock In \bibinfo{booktitle}{\emph{NIPS}}. \bibinfo{pages}{5021--5029}.
\newblock


\bibitem[Fine and Gray(1999)]%
        {fine-gray}
\bibfield{author}{\bibinfo{person}{J.~P. Fine} {and} \bibinfo{person}{R.~J.
  Gray}.} \bibinfo{year}{1999}\natexlab{}.
\newblock \showarticletitle{A Proportional Hazards Model for the
  Subdistribution of a Competing Risk}.
\newblock \bibinfo{journal}{\emph{Journal of the American statistical
  association}} \bibinfo{volume}{94}, \bibinfo{number}{446}
  (\bibinfo{year}{1999}), \bibinfo{pages}{496--509}.
\newblock


\bibitem[Fraccaro et~al\mbox{.}(2017)]%
        {nips17-marco-kvae}
\bibfield{author}{\bibinfo{person}{Marco Fraccaro}, \bibinfo{person}{Simon
  Kamronn}, \bibinfo{person}{Ulrich Paquet}, {and} \bibinfo{person}{Ole
  Winther}.} \bibinfo{year}{2017}\natexlab{}.
\newblock \showarticletitle{A Disentangled Recognition and Nonlinear Dynamics
  Model for Unsupervised Learning}.
\newblock In \bibinfo{booktitle}{\emph{NIPS}}.
\newblock


\bibitem[Fraccaro et~al\mbox{.}(2016)]%
        {nips16-marco-srnn}
\bibfield{author}{\bibinfo{person}{Marco Fraccaro},
  \bibinfo{person}{Søren~Kaae Sønderby}, \bibinfo{person}{Ulrich Paquet},
  {and} \bibinfo{person}{Ole Winther}.} \bibinfo{year}{2016}\natexlab{}.
\newblock \showarticletitle{Sequential Neural Models with Stochastic Layers}.
\newblock In \bibinfo{booktitle}{\emph{NIPS}}.
\newblock


\bibitem[Ghassemi et~al\mbox{.}(2017)]%
        {ghassemi-switching}
\bibfield{author}{\bibinfo{person}{Marzyeh Ghassemi}, \bibinfo{person}{Mike
  Wu}, \bibinfo{person}{Michael~C. Hughes}, \bibinfo{person}{Peter Szolovits},
  {and} \bibinfo{person}{Finale Doshi-Velez}.} \bibinfo{year}{2017}\natexlab{}.
\newblock \showarticletitle{Predicting intervention onset in the ICU with
  switching state space models}.
\newblock \bibinfo{journal}{\emph{American Medical Informatics Association
  (AMIA),}}.
\newblock


\bibitem[Giunchiglia et~al\mbox{.}(2018)]%
        {icann-rnn-surv}
\bibfield{author}{\bibinfo{person}{E. Giunchiglia}, \bibinfo{person}{A.
  Nemchenko}, {and} \bibinfo{person}{M. van~der Schaar}.}
  \bibinfo{year}{2018}\natexlab{}.
\newblock \showarticletitle{RNN-SURV: A Deep Recurrent Model for Survival
  Analysis}. In \bibinfo{booktitle}{\emph{International Conference on
  Artificial Neural Networks (ICANN)}}.
\newblock


\bibitem[Johnson et~al\mbox{.}(2016)]%
        {johnson16mimic}
\bibfield{author}{\bibinfo{person}{Alistair~E.W. Johnson},
  \bibinfo{person}{Tom~J. Pollard}, \bibinfo{person}{Lu Shen},
  \bibinfo{person}{Li wei H.~Lehman}, \bibinfo{person}{Mengling Feng},
  \bibinfo{person}{Mohammad Ghassemi}, \bibinfo{person}{Benjamin Moody},
  \bibinfo{person}{Peter Szolovits}, \bibinfo{person}{Leo~Anthony Celi}, {and}
  \bibinfo{person}{Roger~G. Mark}.} \bibinfo{year}{2016}\natexlab{}.
\newblock \showarticletitle{{MIMIC-III}, A Freely Accessible Critical Care
  Database}.
\newblock \bibinfo{journal}{\emph{Scientific Data}}  \bibinfo{volume}{3}
  (\bibinfo{year}{2016}).
\newblock
\newblock
\shownote{Article number: 160035}.


\bibitem[Jones et~al\mbox{.}(2010)]%
        {sofa_score}
\bibfield{author}{\bibinfo{person}{Alan~E. Jones}, \bibinfo{person}{Stephen
  Trzeciak}, {and} \bibinfo{person}{MD Jeffrey A.~Kline}.}
  \bibinfo{year}{2010}\natexlab{}.
\newblock \showarticletitle{The Sequential Organ Failure Assessment score for
  predicting outcome in patients with severe sepsis and evidence of
  hypoperfusion at the time of emergency department presentation}.
\newblock


\bibitem[Karl et~al\mbox{.}(2017)]%
        {iclr17-dvbf}
\bibfield{author}{\bibinfo{person}{Maximilian Karl},
  \bibinfo{person}{Maximilian Soelch}, \bibinfo{person}{Justin Bayer}, {and}
  \bibinfo{person}{Patrick van~der Smagt}.} \bibinfo{year}{2017}\natexlab{}.
\newblock \showarticletitle{Deep Variational Bayes Filters: Unsupervised
  Learning of State Space Models from Raw Data}.
\newblock In \bibinfo{booktitle}{\emph{ICLR}}.
\newblock


\bibitem[Katzman et~al\mbox{.}(2018)]%
        {bmc18-katzman-deepsurv}
\bibfield{author}{\bibinfo{person}{Jared~L. Katzman}, \bibinfo{person}{Uri
  Shaham}, \bibinfo{person}{Alexander Cloninger}, \bibinfo{person}{Jonathan
  Bates}, \bibinfo{person}{Tingting Jiang}, {and} \bibinfo{person}{Yuval
  Kluger}.} \bibinfo{year}{2018}\natexlab{}.
\newblock \showarticletitle{DeepSurv: personalized treatment recommender system
  using a Cox proportional hazards deep neural network}.
\newblock \bibinfo{journal}{\emph{BMC Medical Research Methodology}}
  \bibinfo{volume}{18}, \bibinfo{number}{1} (\bibinfo{year}{2018}),
  \bibinfo{pages}{24}.
\newblock


\bibitem[Krishnan et~al\mbox{.}(2015)]%
        {nipsw15-dkf}
\bibfield{author}{\bibinfo{person}{Rahul~G. Krishnan}, \bibinfo{person}{Uri
  Shalit}, {and} \bibinfo{person}{David Sontag}.}
  \bibinfo{year}{2015}\natexlab{}.
\newblock \showarticletitle{Deep Kalman Filters}.
\newblock \bibinfo{journal}{\emph{CoRR}}  \bibinfo{volume}{abs/1511.05121}
  (\bibinfo{year}{2015}).
\newblock


\bibitem[Krishnan et~al\mbox{.}(2017)]%
        {aaai17-sontag-dmm}
\bibfield{author}{\bibinfo{person}{Rahul~G Krishnan}, \bibinfo{person}{Uri
  Shalit}, {and} \bibinfo{person}{David Sontag}.}
  \bibinfo{year}{2017}\natexlab{}.
\newblock \showarticletitle{Structured Inference Networks for Nonlinear State
  Space Models}. In \bibinfo{booktitle}{\emph{AAAI}}.
\newblock


\bibitem[Lee et~al\mbox{.}(2018)]%
        {aaai18-schaar-deephit}
\bibfield{author}{\bibinfo{person}{Changhee Lee}, \bibinfo{person}{William~R.
  Zame}, \bibinfo{person}{Jinsung Yoon}, {and} \bibinfo{person}{Mihaela van~der
  Schaar}.} \bibinfo{year}{2018}\natexlab{}.
\newblock \showarticletitle{DeepHit: A Deep Learning Approach to Survival
  Analysis with Competing Risks}.
\newblock In \bibinfo{booktitle}{\emph{AAAI}}.
\newblock


\bibitem[Lipton et~al\mbox{.}(2016)]%
        {iclr16-lipton}
\bibfield{author}{\bibinfo{person}{Zachary~C Lipton}, \bibinfo{person}{David~C
  Kale}, \bibinfo{person}{Charles Elkan}, {and} \bibinfo{person}{Randall
  Wetzel}.} \bibinfo{year}{2016}\natexlab{}.
\newblock \showarticletitle{Learning to Diagnose with {LSTM} Recurrent Neural
  Networks}. In \bibinfo{booktitle}{\emph{International Conference on Learning
  Representations (ICLR)}}.
\newblock


\bibitem[Liu and Hauskrecht(2013)]%
        {aim2013-liu}
\bibfield{author}{\bibinfo{person}{Zitao Liu} {and} \bibinfo{person}{Milos
  Hauskrecht}.} \bibinfo{year}{2013}\natexlab{}.
\newblock \showarticletitle{Clinical Time Series Prediction with a Hierarchical
  Dynamical System}.
\newblock \bibinfo{journal}{\emph{Artificial Intelligence in Medicine}}
  (\bibinfo{year}{2013}), \bibinfo{pages}{227--237}.
\newblock


\bibitem[Liu and Hauskrecht(2016)]%
        {aaai2016-liu}
\bibfield{author}{\bibinfo{person}{Zitao Liu} {and} \bibinfo{person}{Milos
  Hauskrecht}.} \bibinfo{year}{2016}\natexlab{}.
\newblock \showarticletitle{Learning Adaptive Forecasting Models from
  Irregularly Sampled Multivariate Clinical Data}. In
  \bibinfo{booktitle}{\emph{AAAI}}.
\newblock


\bibitem[Lu~Wang and Ye(2017)]%
        {icdm17-li-multitask}
\bibfield{author}{\bibinfo{person}{Jiayu Zhou Dongxiao~Zhu Lu~Wang, Yan~Li}
  {and} \bibinfo{person}{Jieping Ye}.} \bibinfo{year}{2017}\natexlab{}.
\newblock \showarticletitle{Multi-task Survival Analysis}.
\newblock In \bibinfo{booktitle}{\emph{IEEE International Conference on Data
  Mining}}.
\newblock


\bibitem[M et~al\mbox{.}(2017)]%
        {jaim2017-wu}
\bibfield{author}{\bibinfo{person}{Wu M}, \bibinfo{person}{Ghassemi M},
  \bibinfo{person}{Feng M}, \bibinfo{person}{Celi LA},
  \bibinfo{person}{Szolovits P}, {and} \bibinfo{person}{Doshi-Velez F}.}
  \bibinfo{year}{2017}\natexlab{}.
\newblock \showarticletitle{Understanding vasopressor intervention and weaning:
  risk prediction in a public heterogeneous clinical time series database}.
\newblock \bibinfo{journal}{\emph{J Am Med Inform Assoc}}
  (\bibinfo{year}{2017}), \bibinfo{pages}{488--495}.
\newblock


\bibitem[Raghu et~al\mbox{.}(2017)]%
        {pmlr-v68-raghu17a}
\bibfield{author}{\bibinfo{person}{Aniruddh Raghu}, \bibinfo{person}{Matthieu
  Komorowski}, \bibinfo{person}{Leo~Anthony Celi}, \bibinfo{person}{Peter
  Szolovits}, {and} \bibinfo{person}{Marzyeh Ghassemi}.}
  \bibinfo{year}{2017}\natexlab{}.
\newblock \showarticletitle{Continuous State-Space Models for Optimal Sepsis
  Treatment: a Deep Reinforcement Learning Approach}. In
  \bibinfo{booktitle}{\emph{Proceedings of the 2nd Machine Learning for
  Healthcare Conference}}. \bibinfo{pages}{147--163}.
\newblock


\bibitem[Rajkomar and et~al.(2018)]%
        {rajkomar18scalable}
\bibfield{author}{\bibinfo{person}{Alvin Rajkomar} {and} \bibinfo{person}{et
  al.}} \bibinfo{year}{2018}\natexlab{}.
\newblock \showarticletitle{Scalable and Accurate Deep Learning with Electronic
  Health Records}.
\newblock \bibinfo{journal}{\emph{Digital Medicine}}  \bibinfo{volume}{1}
  (\bibinfo{year}{2018}).
\newblock
\newblock
\shownote{Article number: 18}.


\bibitem[Ranganath et~al\mbox{.}(2016)]%
        {pmlr16-blei-dsa}
\bibfield{author}{\bibinfo{person}{Rajesh Ranganath}, \bibinfo{person}{Adler
  Perotte}, \bibinfo{person}{Noémie Elhadad}, {and} \bibinfo{person}{David
  Blei}.} \bibinfo{year}{2016}\natexlab{}.
\newblock \showarticletitle{Deep Survival Analysis}.
\newblock In \bibinfo{booktitle}{\emph{Proceedings of Machine Learning
  Research}}. Vol.~\bibinfo{volume}{56}. \bibinfo{pages}{101--114}.
\newblock


\bibitem[Ren et~al\mbox{.}(2019)]%
        {aaai19-ren}
\bibfield{author}{\bibinfo{person}{Kan Ren}, \bibinfo{person}{Jiarui Qin},
  \bibinfo{person}{Lei Zheng}, \bibinfo{person}{Zhengyu Yang},
  \bibinfo{person}{Weinan Zhang}, \bibinfo{person}{Lin Qiu}, {and}
  \bibinfo{person}{Yong Yu}.} \bibinfo{year}{2019}\natexlab{}.
\newblock \showarticletitle{Deep Recurrent Survival Analysis}. In
  \bibinfo{booktitle}{\emph{AAAI}}.
\newblock


\bibitem[Sha and Wang(2017)]%
        {bcb2017-sha}
\bibfield{author}{\bibinfo{person}{Ying Sha} {and} \bibinfo{person}{May~D
  Wang}.} \bibinfo{year}{2017}\natexlab{}.
\newblock \showarticletitle{Interpretable Predictions of Clinical Outcomes with
  An Attention-based Recurrent Neural Network}. In
  \bibinfo{booktitle}{\emph{Proceedings of the 8th {ACM} International
  Conference on Bioinformatics, Computational Biology,and Health Informatics}}.
\newblock


\bibitem[Song et~al\mbox{.}(2018)]%
        {aaai2018-song}
\bibfield{author}{\bibinfo{person}{Huan Song}, \bibinfo{person}{Deepta Rajan},
  \bibinfo{person}{Jayaraman~J Thiagarajan}, {and} \bibinfo{person}{Andreas
  Spanias}.} \bibinfo{year}{2018}\natexlab{}.
\newblock \showarticletitle{Attend and Diagnose: Clinical Time Series Analysis
  Using Attention Models}. In \bibinfo{booktitle}{\emph{AAAI}}.
\newblock


\bibitem[Xue et~al\mbox{.}(2019)]%
        {nipsw19-xue-dssm}
\bibfield{author}{\bibinfo{person}{Yuan Xue}, \bibinfo{person}{Denny Zhou},
  \bibinfo{person}{Nan Du}, \bibinfo{person}{Andrew~M. Dai},
  \bibinfo{person}{Zhen Xu}, \bibinfo{person}{Kun Zhang}, {and}
  \bibinfo{person}{Claire Cui}.} \bibinfo{year}{2019}\natexlab{}.
\newblock \showarticletitle{Deep Physiological State Space Model for Clinical
  Forecasting}. In \bibinfo{booktitle}{\emph{NeurIPS Workshop on Machine
  Learning for Health}}.
\newblock


\bibitem[Yan~Li and Reddy(2016)]%
        {kdd16-li-multitask}
\bibfield{author}{\bibinfo{person}{Jieping~Ye Yan~Li, Jie~Wang} {and}
  \bibinfo{person}{Chandan~K. Reddy}.} \bibinfo{year}{2016}\natexlab{}.
\newblock \showarticletitle{A Multi-Task Learning Formulation for Survival
  Analysis}.
\newblock In \bibinfo{booktitle}{\emph{22nd ACM SIGKDD}}.
\newblock


\bibitem[Zhang and Zhou(2018)]%
        {nips18-zhang-lomax}
\bibfield{author}{\bibinfo{person}{Quan Zhang} {and} \bibinfo{person}{Mingyuan
  Zhou}.} \bibinfo{year}{2018}\natexlab{}.
\newblock \showarticletitle{Nonparametric Bayesian Lomax delegate racing for
  survival analysis with competing risks}.
\newblock In \bibinfo{booktitle}{\emph{NeurIPS}}.
\newblock


\end{thebibliography}

\newpage
\section{Supplementary Materials}
\label{appendix:preproc}

\subsection{Data Preprocessing Details}
We preprocess the MIMIC-III dataset in the following steps.

\begin{enumerate}
    \item {\bf Code harmonization}. This is a manual process based on the inputs from clinical experts. In this step, the medical codes corresponding to the same measurements from different coding systems, including LONIC and MIMIC specific coding, are harmonized into the same entity. For example, serum creatinine is associated the following MIMIC-III specific codes: 220615, 50912, 1525, 3750, 791, and LONIC code 2160-0. 
    \item {\bf Unit conversion}. This is an automated process with manual review. In MIMIC-III, a medical code may be used in multiple units and sometimes miss a unit. To determine whether these two entries correspond to the same measurement concept, we derive its value range and mean under different units and test whether they are similar to each other. The final results are reviewed manually. 
    \item {\bf Outlier removal}. We derive the distribution of each measurement code after harmonization and unit conversion. We remove the outliers, defined as below $0.1 \times$ the value at $1$ percentile or above $10 \times$ the value at $99$ percentile. 
    \item {\bf Value normalization}. We collect the mean and standard deviation over the cleaned dataset for each harmonized code and compute its z-score as feature value. 
\end{enumerate}

\subsection{Features}

We record the features and their units used in the experiment in Table~\ref{table:feature-1}-\ref{table:feature-3}. 

\begin{table}[h!]
\centering
\footnotesize
\begin{tabular}{l l} \\
\toprule 
%Feature & unit & MIMIC code. & LONIC code & Num \\
Feature & unit \\
access pressure	&	MMHG	\\
albumin	&	G PER DL	\\
alt	&	IU PER L	\\
anion gap	&	MEQ PER L	\\
ap	&	IU PER L	\\
arterial base excess	&	MEQ PER L	\\
arterial bicarbonate	&	MEQ PER L	\\
arterial pco2	&	MMHG	\\
arterialph	&	PH	\\
arterial po2	&	MMHG	\\
ast	&	IU PER L	\\
base excess	&	MEQ PER L	\\
basophils	&	PERCENT	\\
blood flow	&	ML PER MIN	\\
bp diastolic invasive	&	MMHG	\\
bp diastolic non invasive	&	MMHG	\\
bp map invasive	&	MMHG	\\
bp mean non invasive	&	MMHG	\\
bp systolic invasive	&	MMHG	\\
bp systolic non invasive	&	MMHG	\\
bun	&	MG PER DL	\\
calcium	&		\\
calcium	&	MEQ PER L	\\
\bottomrule
\end{tabular}
\caption{\small  Observation Feature and Unit  (Part 1)}
\label{table:feature-1}
\end{table}

\begin{table}[h!]
\centering
\footnotesize
\begin{tabular}{l l} \\
\toprule 
%Feature & unit & MIMIC code. & LONIC code & Num \\
Feature & unit \\

cardiac index	&	UNKNOWN UOM	\\
cardiac output rate	&	L PER MIN	\\
chloride	&	MEQ PER L	\\
co2	&	MEQ PER L	\\

creatine kinase	&	IU PER L	\\
creatinine	&	MG PER DL	\\
creatinine kinase mb	&	NG PER ML	\\
cvp	&	MMHG	\\
eosinophils	&	PERCENT	\\
exhaled minute ventilation low	&		\\
exhaled minute ventilation low	&	L PER MIN	\\
expiratory ratio	&	RATIO	\\
fio2 analyzed	&	TORR	\\
glucose	&	MG PER DL	\\
glucose poc	&	MG PER DL	\\
heart rate	&	BPM	\\
hematocrit	&	PERCENT	\\
hemoglobin	&	G PER DL	\\
inr	&	RATIO	\\
inr	&		\\
insp pressure	&	CM H2O	\\
insp time	&	S	\\
inspiratory ratio	&	RATIO	\\
ionized calcium	&	MEQ PER L	\\
ketones urine	&	MG PER DL	\\
lactate	&	MMOL PER L	\\
lymphocytes diff	&	PERCENT	\\
magnesium	&	MG PER DL	\\
mean ch	&	PG	\\
mean chc	&	PERCENT	\\
mean cv	&	FL	\\
minute ventilation obs	&	L PER MIN	\\
monocytes	&	PERCENT	\\
neutrophils urine	&	PERCENT	\\
o2 flow	&	L PER MIN	\\
o2 saturation	&	PERCENT	\\
o2 saturation p	&	PERCENT	\\
p co2	&	MMHG	\\
p o2	&	MMHG	\\
paw	&	CM H2O	\\
peep observed	&	CM H2O	\\
ph	&	PH	\\
ph urine	&	PH	\\
phosphorous	&	MEQ PER L	\\
plateau pressure	&	CM H2O	\\
platelet	&		\\
platelet	&	K PER UL	\\
potassium	&	MEQ PER L	\\
potassium urine	&	MEQ PER L	\\
protein urine	&	MG PER DL	\\
psv	&	UNKNOWN UOM	\\
pt	&	S	\\
ptt	&	S	\\
rbc	&	PER UL	\\
rdw	&	PERCENT	\\
replacement rate	&	ML PER H	\\
respiratory rate	&	BREATHS PER MIN	\\
respiratory rate spont	&	BREATHS PER MIN	\\
respiratory rate total	&	BPM	\\
rrt output	&	ML	\\
sodium	&	MEQ PER L	\\
specific gravity urine	&		\\
svo2	&	PERCENT	\\
tbili	&	MG PER DL	\\
tco2	&	MEQ PER L	\\
temperature	&	CEL	\\
troponin t	&	NG PER ML	\\
urine output	&	ML	\\
urine output foley	&	ML	\\
vt obs	&	ML PER BREATH	\\
vt spont	&	ML PER BREATH	\\
wbc count	&	K PER UL	\\
weight	&	KG	\\
\bottomrule
\end{tabular}
\caption{\small Observation Feature and Unit (Part 2)}
\label{table:feature-2}
\end{table}

\begin{table}[h!]
\centering
\footnotesize
\begin{tabular}{l l} \\
\toprule 
%Feature & unit & MIMIC code. & LONIC code & Num \\
Feature & unit \\

30042	&	MCG KG MIN	\\
30043	&	MCG KG MIN	\\
30044	&	MCG MIN	\\
30047	&	MCG MIN	\\
30120	&	MCG KG MIN	\\
30127	&	MCG MIN	\\
30306	&	MCG KG MIN	\\
30307	&	MCG KG MIN	\\
	&		\\
dialysate rate	&	ML PER H	\\
fi o2	&	PERCENT	\\
peep	&	CM H2O	\\
pip	&	CM H2O	\\
respiratory rate setting	&	BPM	\\
vt set	&	ML PER BREATH	\\ 

\bottomrule
\end{tabular}
\caption{\small Intervention Feature and Unit}
\label{table:feature-3}
\end{table}

\end{document}